%% file: ejemplo_espanol.tex
\documentclass[a4paper]{llncs}

\usepackage[colorlinks=true,linkcolor=blue,citecolor=blue]{hyperref}
\usepackage{times}
\usepackage{graphicx,color}
\usepackage[T1]{fontenc}
\usepackage{verbatim}
\usepackage{longtable}
\usepackage{tabularx} 
\usepackage{multirow}
\usepackage{multicol}
\usepackage{float}
\usepackage{listings}
\usepackage{anysize}
\usepackage{geometry}
\usepackage{array}
\usepackage{hhline}
\usepackage{latexsym}
\usepackage{amsmath}
\usepackage{amssymb}
\usepackage{algorithm}
\usepackage{algorithmic}
\usepackage{booktabs}
\usepackage{colortbl}
\usepackage{listings}
\usepackage{setspace}
\usepackage{ragged2e}
\usepackage{tabularray}
\usepackage{color}
\usepackage{makecell}

\usepackage[]{inputenc}

\begin{document}
\title{Sólo Escúchame: Spanish Emotional Accompaniment Chatbot}
\titlerunning{Sólo Escúchame}
%


\author{Bruno Gil Ram\'irez$^{\ast}$, Jessica L\'opez Espejel$^{\dagger}$, \\Mar\'ia del Carmen Santiago D\'iaz$^{\ast}$, Gustavo Trinidad Rub\'in Linares$^{\ast}$}

\authorrunning{Gil Ram\'irez et al.}
\institute{$^{\ast}$ Benem\'erita Universidad Aut\'onoma de Puebla, Mexico \\ $^{\dagger}$ Novelis Research and Innovation Lab, France
\\
}

\maketitle              

\begin{abstract}
    According to the World Health Organization (WHO), suicide was the fourth leading cause of death in the world for individuals aged $15$ to $29$ in $2019$. Given the rapid increase in mental health issues, providing psychological support is both crucial and urgent. In this paper: ($1$) we propose \textit{Sólo Escúchame}, the first open-source Spanish emotional assistance chatbot, based on LLaMA-2-7b-Chat. ($2$) We introduced the HEAR (Hispanic Emotional Accompaniment Responses) dataset, compiled from multiple English sources translated into Spanish, as well as generic data generated using ChatGPT-3.5-Turbo. Finally, ($3$) we propose an evaluation metric based on two semi-automatic assessment methods. Our system outperforms a range of state-of-the-art models in providing psychological assistance in Spanish. Our models and datasets are publicly available to facilitate reproducibility.

\end{abstract}

\section{Introduction}

    Research on conversational chatbots for mental health has grown significantly in recent years~\cite{Darcy2023_Woebot,Inkster2023_wysa,Raws2023_tess}. These chatbots offer a promising avenue to address the rising mental health concerns, particularly in the wake of the COVID-19 pandemic~\cite{Boettler2020_covid,ciotti2020_covid}. The pandemic's drastic impact on routines~\cite{Naglaa2020_Antivirus}, including the rise of remote work and home schooling, has been linked to increased rates of anxiety, depression, and even suicide among individuals aged $15$ to $29$ globally~\cite{Asarnow2019_suicide,Flaviane2021_depression,lee2020_incremental,mazza2020_anxiety,oppenheimer2020_neural}. However, existing resources to combat these issues are often limited or closed-source, hindering their effectiveness~\cite{Shah2020_FocusOM}.

    Despite the impressive capabilities of current Language Models (LLMs) such as Chinchilla~\cite{Hoffmann2022_chinchilla}, PaLM~\cite{Chowdhery2022_PaLMSL}, LLaMA~\cite{Touvron2023_LLaMa1,Touvron2023_LLaMa2}, ChatGPT~\cite{chatGPT3.5,chatGPT4}, BARD~\cite{manyika2023_BARD}, Mistral~\cite{jiang2023_mistral}, and Gemini~\cite{geminiteam2023_gemini}, it is noteworthy to mention that they have not been explicitly designed or optimized for tasks related to emotional support. Furthermore, the predominant focus of these models on the English language presents a significant barrier to non-English speakers, limiting their accessibility and effectiveness in providing emotional support to individuals speaking different languages.

    To bridge these gaps, we propose ``\textit{Sólo Escúchame} (Just Listen to Me)'', an open-source Spanish emotional assistance language model. Inspired by similar psychological support chatbots~\cite{Narynov2021_Chatbot,Salhi2021_chatbot}, \textit{Sólo Escúchame} aims to provide accessible support, particularly for those who may not have immediate access to a professional psychologist due to personal or financial limitations. It is important to emphasize that \textit{Sólo Escúchame} serves as a supplementary tool or a resource for psychologists, and does not replace professional mental healthcare.
    
    Our contributions include:
     
    $\bullet$ We introduce HEAR (Hispanic Emotional Accompaniment Responses) dataset, specialized in emotional accompaniment. Our dataset is publicly available to facilitate future research.
       
    $\bullet$ We propose \textit{``Sólo Escúchame''}, a chatbot trained on our HEAR dataset and runs efficiently on CPUs. To the best of our knowledge, it is the first open-source Spanish chatbot designed for psychological assistance.
    
    $\bullet$ We designed efficient prompts for ($1$) generating generic data using GPT-3.5-Turbo, and ($2$) training the \textit{Sólo Escúchame} model through instruction-following demonstrations.
            
    $\bullet$ We introduce semi-manual evaluation metrics for fair model comparison. 

\section{Related Work}

    \paragraph{\underline{Psychological support chatbots.}} Due to the COVID-19 pandemic, numerous chatbot solutions have emerged to provide psychological support. Chatbots have proven effective in reducing symptoms of depression and anxiety~\cite{kovsecka2023_Robot—Chatbot,may2022_security}. These chatbots are often available as mobile applications, such as Woebot~\cite{Darcy2023_Woebot}, Wysa~\cite{Inkster2023_wysa}, Tess~\cite{Raws2023_tess} and Youper~\cite{Hamilton2024_youper}. Woebot Health offers an enterprise solution that improves access to mental health support by enhancing emotional regulation skills and aiding in mood monitoring and management. Wysa AI Coach is an AI-driven service that provides emotionally intelligent responses to users' emotions, assisting individuals with low mood, stress, or anxiety. Tess, another mental health chatbot, is created to help individuals dealing with panic attacks or those who need to discuss their thoughts before sleeping. At last, Youper chatbot app utilizes cognitive behavioral therapy (CBT) and positive psychology to aid users in managing anxiety and depression.

    It is worth noting that earlier emotional chatbots for healthcare were primarily designed for English speakers, as most available datasets in this field are in English~\cite{Lambert2024SelfDirected,li2017_dailydialog,Rajpurkar2016_SQuAD,Taori2021_alpaca}. Thus, separate chatbots were developed for the Spanish-speaking community, although these remain very few. Among these, Perla~\cite{arrabales2020_perla} is a conversational agent specialized in conducting interviews based on the Patient Health Questionnaire-9 (PHQ-9). Similarly,~\cite{Yanes2022_sentiment} developed a sentiment analysis chatbot designed to detect potential signs of mental health issues, intended for high school students typically aged $14$ to $18$ years old. Unfortunately, none of the prior Spanish language chatbots are open source.

    Fortuitously, recent years have seen the emergence of open-source models~\cite{jiang2023_mistral,Touvron2023_LLaMa2,wu2024laminilm}, enabling free user customization and broadening access to advanced tools for mental health support. Moreover, these models have facilitated the development of more inclusive and versatile chatbots tailored to diverse linguistic and cultural needs. Next, we showcase some of the most powerful chatbots from both open-source and proprietary models, highlighting the enhancements achieved through the instruction-following demonstrations approach.

    \paragraph{\underline{Large Language Models (LLMs).}} The development of chatbots reached a significant breakthrough with the introduction of the instruct-following demonstration approach~\cite{Ouyang2022_instructions}. This method aligns language models (LMs) with user intent, enabling chatbots to understand and execute specific instructions or commands from users. The significance of this approach became widely recognized with the release of ChatGPT-3.5 in 2022~\cite{chatGPT3.5}.
    
    ChatGPT-3.5, with its 175 billion parameters, marked the beginning of a new era in chatbot research. This era has seen the emergence of powerful and close-source models such as BARD~\cite{manyika2023_BARD}, Gemini-Ultra~\cite{geminiteam2023_gemini} (1.56 trillion parameters), Claude~\cite{claude_chatbot}, and ChatGPT-4~\cite{chatGPT4} (1.7 trillion parameters). However, the large number of parameters in these models often requires substantial computational resources, making them difficult for small businesses and individuals to access.

    Consequently, some research has shifted its focus from solely generating accurate responses to developing smaller, more accessible models~\cite{Hoffmann2022_chinchilla}, as well as creating open-source models~\cite{jiang2023_mistral,Taori2021_alpaca,Touvron2023_LLaMa2}. Notable examples include LLaMA-1-2~\cite{Touvron2023_LLaMa1,Touvron2023_LLaMa2} (up to 70 billion parameters); Vicuna~\cite{Taori2021_alpaca}, an open-source chatbot with 13 billion parameters, fine-tuned from LLaMA;  Mistral~\cite{jiang2023_mistral}, based on a transformer architecture~\cite{Vaswani2017_transformers} with 7 billion parameters; and Mixtral8x7B~\cite{jiang2024_mixtral}, a sparse mixture of experts model. For our work, we have selected the LLaMA-2-7b-Chat model. This choice is based on two critical factors: it is an open-source model and, at the time of our experiments, it represents the state-of-the-art in its category.

    \section{\textit{Sólo Escúchame} Language Model}

    \subsection{Dataset}
    
    This is one of our main contributions and it is made of two stages. The first one is to create an Emotion Recognition Dataset where each sample $X=\{x, y\}$ contains a text $x$ and a class $y \in \{$\textit{affection}, \textit{happiness}, \textit{admiration}, \textit{anger}, \textit{sadness}, \textit{optimism}, \textit{hate}, \textit{surprise}, \textit{fear}, \textit{calm}, \textit{disgust}$\}$. The second stage takes the first dataset and creates the final dataset named HEAR, where each sample contains a text, its class, and a generic text generated by GPT-3.5-Turbo. We detail the two stages below:

    \paragraph{\underline{Stage 1 : Create the Hispanic Emotion Recognition Based on Plutchik's Wheel (HRECPW) Dataset.}}
    
    To build our Spanish dataset for emotion recognition~\footnote{\url{https://huggingface.co/datasets/BrunoGR/HRECPW-Hispanic_Responses_for_Emotional_Classification_based_on_Plutchik_Wheel}}, we leveraged diverse English sources, including TweetEval~\cite{barbieri-etal-2020-tweeteval}, DailyDialog~\cite{li2017_dailydialog}, HappyDB~\cite{asai2018_happydb}, and responses from $72$ surveys we conducted with various individuals to capture a range of emotional examples and contexts. First, we preprocessed the dataset by  removing all personal information and web links. At this stage, the dataset contained $13$ classes of emotions: \textit{affection, achievement, joy, optimism, calm, anger, disgust, fear, sadness, surprise, love, hate and offensive}. Second, we translated the datasets from English to Spanish and re-annotated the data using Plutchik's Wheel of Emotions~\cite{plutchik1980_general}. This led to merging some classes (\textit{love} into \textit{affection}, \textit{offensive} into \textit{anger}) and renaming others (\textit{achievement} to \textit{admiration}) for better alignment with the expressed emotions, resulting in a final set of $11$ classes: $8$ principal emotions (\textit{admiration, anger, disgust, fear, hate, joy, sadness, surprise}) and $3$ compound emotions (\textit{calm, optimism, affection}). Refer to Table~\ref{tab:examples_emotion_recognition} for examples of dataset samples.

    \begin{table}[!ht]
        \centering
        \begin{tabular}{|p{0.8\linewidth}|p{0.2\linewidth}|} 
            \hline
            \textbf{Text} & \textbf{Label} \\
            \hline
            \small
            Every negative experience is an opportunity to grow and learn. 
            &
            \small
            Optimism 
            \\\hline
            \small
            I found a recipe that will make cooking chicken noodle soup easy for a class I am teaching 
            & 
            \small
            Admiration
            \\\hline
            \small
            It bothers me that you always behave aggressively and cannot have a civilized conversation 
            &
            \small
            Anger
            \\\hline
        \end{tabular}
        \vspace{-0.5em}
        \caption{Some samples of Emotion Recognition dataset. \textbf{Note that the samples in our dataset are in Spanish.}}
        \label{tab:examples_emotion_recognition}
    \end{table}

    The emotion recognition dataset was notably imbalanced, with a significant surplus of the \textit{affection} class, which contained $32,837$ samples. In contrast, the \textit{disgust} class had the fewest samples, totaling only $303$. To address this issue, we employed under-sampling for the over-represented classes by randomly eliminating samples. This process reduced each class to a uniform size of $11,000$ samples. This number per class was acceptable size to this dataset, and it was permissible to generate between $2,000$ and $10,700$ samples to over-sampling the less-represented classes. At the end of this stage, we generated $48,500$ generic samples for less-represented emotion sets, using the \textit{GPT-3.5-Turbo} model~\cite{chatGPT3.5}. Consequently, each of the $11$ classes contains $11,000$ samples in the training set, $200$ samples in the validation set, and $120$ samples in the test set. In total, this dataset contains $121,000$ examples for training, $2,200$ for validation and $1,320$ for testing.

\begin{figure*}[!ht]
            \centering
            \includegraphics[width=0.65\textwidth]{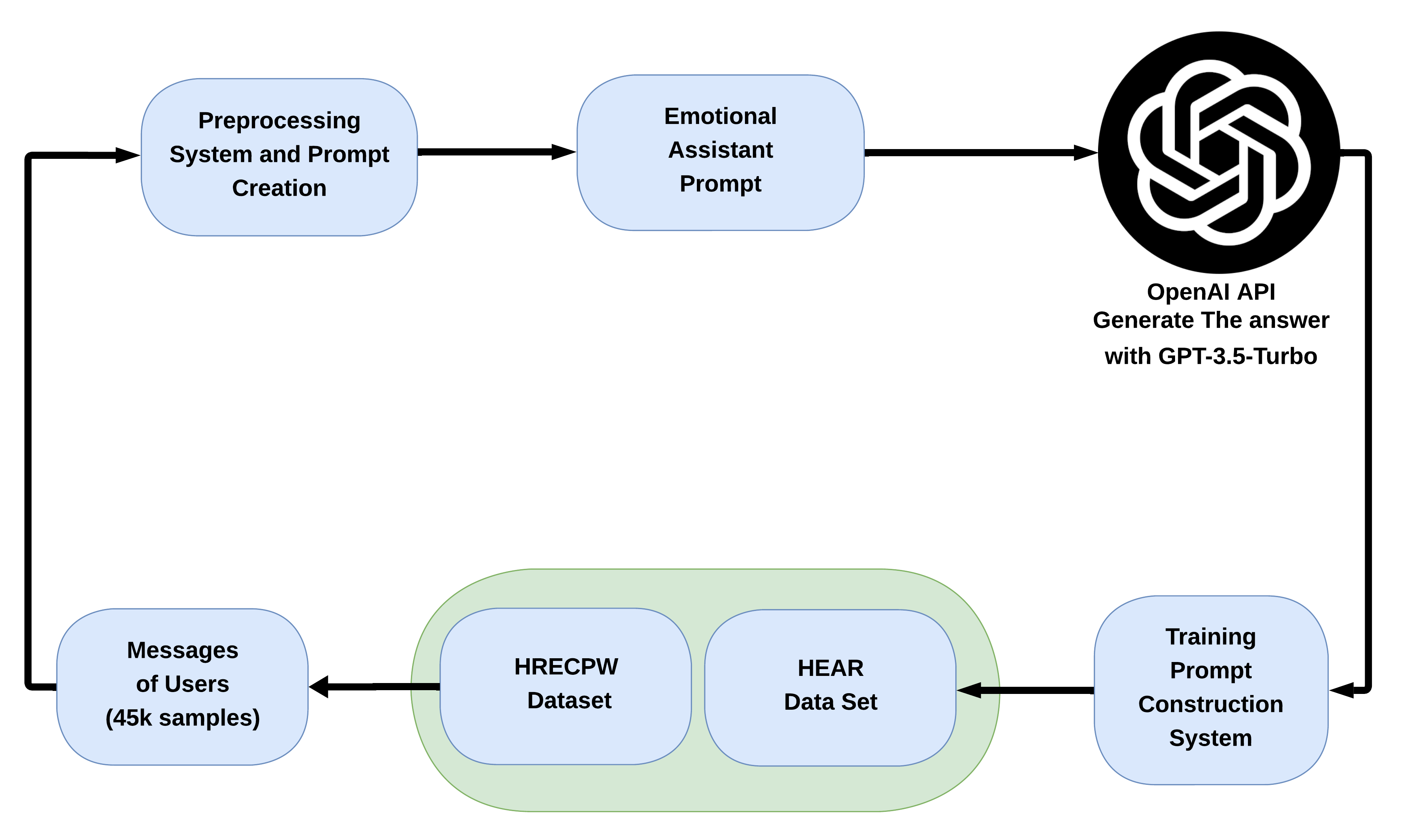}
            \caption{Overview of the procedure to generate generic data using GPT-3.5.}
            \label{fig:overview_generic_data}
    \end{figure*}

    \paragraph{\underline{Stage 2: Create the Hispanic Emotional Accompaniment Responses (HEAR) Dataset.}}  After balancing the dataset for emotion recognition across the $11$ classes, we randomly extracted $3,771$ samples for each emotion class from the training set, and kept the entire validation and test sets. Therefore, the final dataset comprises $41,481$ training, $2,200$ validation, and $1,320$ test samples, respectively. 
    We named this dataset HEAR~\footnote{\url{https://huggingface.co/datasets/BrunoGR/HEAR-Hispanic_Emotional_Accompaniment_Responses}}, which stands for Hispanic Emotional Accompaniment Responses dataset. We built it by generating generic responses to each user's symptoms using GPT-3.5-Turbo, as illustrated in Figure~\ref{fig:overview_generic_data}. Table~\ref{tab:text_response_sample} displays  samples from the dataset. This finalized dataset was used to train the model for generating empathetic and suitable responses in emotionally supportive conversations.  

    \begin{table}[!ht]
            \centering
            \begin{tabular}{|p{0.4\linewidth}|p{0.6\linewidth}|}
                \hline
                 \textbf{Input Text} & \textbf{Generic response} \\
                 \hline
                 \small
                  I was pleasantly surprised when I received a thank-you letter for my volunteer work.
                  & 
                  \small
                  What a beautiful surprise to receive a thank-you letter for your volunteer work! That shows how valuable you are to others. Tell me more about that letter; what did it say?  \\ \hline
                  \small
                  It irritates me when people do not respect my privacy on social media.
                  &
                  \small
                  I'm sorry that some people do not respect your privacy on social media. Privacy is a fundamental right. Would you like to discuss more about your concerns and how you would like to protect your privacy online?
                  \\ \hline
            \end{tabular}
            \caption{Samples of HEAR dataset. \textbf{Note that the samples in our dataset are in Spanish.}}
            \label{tab:text_response_sample}
        \end{table}

    \subsection{Model}

    As previously stated, \textit{Sólo Escúchame}~\footnote{\url{https://github.com/BrunoGilRamirez/Just_HEAR_ME}}~\footnote{\url{https://huggingface.co/BrunoGR/Just_HEAR_Me}} is a fine-tuned version of LLaMA-2-7b-Chat, a model proposed by META \cite{Touvron2023_LLaMa2}. One of the main novelties of this LLM is the improvement of the context length using Rotary Positional Embedding (RoPE)~\cite{su2023_rope}, and the use of Grouped-Query Attention (GQA)~\cite{ainslie2023_gqa}.

    We chose this open-source lightweight model for its balanced performance and speed. It is important for our model to run on a CPU to ensure accessibility to a wider audience. Therefore, we include quantized versions of our model with $2$, $4$, and $8$ bits. To perform quantization, we used scripts provided by the llama.cpp project~\cite{Gerganov2023_llamaCPP}.

    \paragraph{\textbf{Fine-Tuning.}} We fine-tuned LLaMA-2-7b-Chat on our HEAR dataset using LoRA (Low Rank Adaptation)~\footnote{\url{https://huggingface.co/BrunoGR/JUST_HEAR_ME-PEFT_Adapter}}\cite{hu2021lora}. LoRA consists in freezing the pre-trained Transformers weights, and only training small rank decomposition matrices instead, thus reducing considerably the total number of trainable parameters for downstream tasks. We utilized Hugging Face's Transformers library to train our model. Through extensive experimentation, we determined that the optimal parameters for our task are $batch~size=15$, $micro~batch~size=5$, $warmup~steps=300$, $learning~rate=5e^{-5}$. For the LoRA technique, we used the following key parameter values: $R=64$, $alpha=128$, $dropout=0.1$.

    \paragraph{\textbf{Prompting.}} We follow the methodology introduced by~\cite{Ouyang2022_instructions} and the Alpaca model~\cite{Taori2021_alpaca} to fine-tune our model using instruction-following demonstrations. Table \ref{tab:training_prompt} shows the prompt we propose to train the model:
    
    \begin{table}[!ht]
        \centering
        \begin{tabular}{|p{1.0\linewidth}|}
            \hline
            \small
             \textit{Below is an instruction that describes a task, paired with an input that provides further context. }

             \textit{\#\#\# instruction: You are an emotional assistant, respond in Spanish in a respectful and appropriate way to the user's emotional situation. If the user appears sad or upset, the assistant should respond empathetically and offer words of encouragement. }

             \textit{\#\#\# input: } input\_text

             \textit{\#\#\# response: } response\_text </s>
             \\
             \hline
        \end{tabular}
        \caption{The proposed training prompt}
        \label{tab:training_prompt}
    \end{table}

    \section{Experiments}
    \label{sec:evaluation_metrics}

    \subsection{Evaluation Protocol} 

    To evaluate the performance of our model \textit{Sólo Escúchame}, we used the entire test set of the HEAR dataset, which contains $1,320$ samples. The evaluation follows two sets of criteria: the active listening technique~\cite{rogers2015active} and the Socratic method~\cite{oblitas-2008}. We selected these criteria because they are widely used by psychologists specializing in cognitive behavioral therapy~\cite{castillero-mimenza-2024}. Our chatbot's goal is to make users feel heard and understood, and both methodologies include measures for these outcomes. 
    
    \textbf{Note that in the next two subsections, "$\rightarrow$" means "answers the question"}.

    \paragraph{$\bullet$~~\underline{\textbf{Active Listening}}:} defined by Carl Rogers~\cite{rogers2015active}, it involves attentive listening to the full conversation, considering verbal and non-verbal cues, and expressing personal feelings. Some aspects, like non-verbal signs, are not evaluable in language models. Therefore, our assessment focuses on the following specific points for our model and other LMs:

    $\circ$ \textit{Contextual Attention} $\rightarrow$ Does the model demonstrate coherence in its responses, showing that it pays attention to the information provided earlier in the conversation?
    
    $\circ$ \textit{Formulation of Clarifying Questions} $\rightarrow$ Does the model ask questions to obtain more details or clarifications about what has been said?

    $\circ$ \textit{Delving into the Conversation} $\rightarrow$ Does the model encourage a deeper conversation while remaining consistent with the context provided by the interlocutor?
        
    $\circ$  \textit{Absence of Judgment or Critique} $\rightarrow$ Does the model avoid expressing judgments or criticisms towards the interlocutor's statements?
    
   $\circ$  \textit{Demonstration of Empathy} $\rightarrow$ (1) Does the model empathetically respond, recognizing and reflecting the emotions expressed by the interlocutor? (2) Does the model acknowledge the emotions expressed by the interlocutor in its response?

   \paragraph{$\bullet$~~\underline{\textbf{Socratic Method:}}} defined by the psychologist~\cite{oblitas-2008}, it involves therapists guiding patients to formulate solutions through reflective questioning. Similar to the active listening technique, there are evaluation criteria that we cannot consider in the assessment of LMs. Consequently, we focus on the following five criteria.
  
    $\circ$  \textit{Use of Inductive Questions} $\rightarrow$ Does the model formulate questions that encourage reflection and critical analysis rather than providing direct answers?

    $\circ$  \textit{Non-Imposition of Ideas} $\rightarrow$ Does the model refrain from imposing ideas or solutions, by guiding the interlocutor to discover their own answers?
    
    $\circ$  \textit{Knowledge Expansion and Construction} $\rightarrow$ Does the model engage in knowledge construction through reflection and continuous dialogue?
            
    $\circ$  \textit{Generation of Cognitive Dissonance} $\rightarrow$ Does the model use dialogue to provoke cognitive dissonance, challenging the beliefs or ideas of the interlocutor?
        
    $\circ$  \textit{Guided Discovery} $\rightarrow$ Does the model primarily employ systematic questioning, guiding the interlocutor to discover their own beliefs and knowledge?

    \vspace{0.5cm}
    The Socratic method and active listening are versatile tools that anyone can use, regardless of their training or experience. These techniques encourage dialogue, reflection, and deep understanding, making them suitable for diverse contexts such as education, therapy, and conflict resolution. They highlight the significance of ensuring that individuals feel acknowledged and comprehended. Because these methods do not require extensive specialized knowledge, they are perfect for our model, which is designed to complement, not replace, a psychologist. By utilizing these approaches, we can evaluate the model's effectiveness based on its impact on users' emotional well-being, ensuring they feel genuinely supported.

    \paragraph{\underline{\textbf{Evaluation Metrics}.}}

    Our evaluation consists of two stages. Firstly, we employ the GPT-4-instruct model~\cite{chatGPT4} to assess user conversations using a crafted prompt covering all evaluation criteria. GPT-4 provides feedback in JSON format, indicating if the conversations meet the criteria. Secondly, manual evaluation verifies GPT-4 assessments. This manual evaluation involves two human annotators: a computer science engineer, and a psychologist with over five years of experience in cognitive behavioral therapy. 

    Both methodologies assign $20\%$ weight to each of the five criteria, contributing to a total score ranging from $0\%$ to $100\%$. The value for each criterion is computed using Equation~\ref{eq:value_criterion}.

    \begin{equation}
        c_x =\frac{2}{n}\sum_{i=0}^{n}Element_i
        \label{eq:value_criterion}
    \end{equation}

     where  $x$ is the criterion being evaluated,  $n$ is the number of elements, and  $Element_i$ is the value of the criterion for the $i^{th}$ element. 

    The overall score is computed as shown in Equation~\ref{eq:overall_score}.

    \begin{equation}
        score = \sum_{x=0}^{m}c_x
        \label{eq:overall_score}
    \end{equation}

    where $c_x$ is the value of the criterion $x$, and  $m$ is the criterion to add.

    \subsection{Compared Models}

    To evaluate the performance of our model, \textit{Sólo Escúchame}, we compare it with the following state-of-the-art models:
    
    $\bullet$ \textbf{LLaMA-2-7b-Chat -} Created by META~\cite{Touvron2023_LLaMa2}, it demonstrates remarkable performance with just $7$ billion parameters. We used it without further fine-tuning.
    
    $\bullet$ \textbf{GPT-3.5 -} Developed by OpenAI, this model stands as a significant milestone in the evolution of chatbot development~\cite{chatGPT3.5}. We used it without further fine-tuning.
    
    $\bullet$ \textbf{GPT2-124M -} This GPT-2 model with $124$ million parameters is a LaMINI version~\cite{Minghao2023_lamini}. We fine-tuned it with the HEAR dataset.
    
    $\bullet$ \textbf{Mixtral 8x7b -} Sparse Mixture of Experts (SMoE) LM~\cite{jiang2024_mixtral} with $7$ billion parameters. We used it without  further fine-tuning.

    \section{Results and Discussion}
    
    Results in Table~\ref{tab:results} show that \textit{Sólo Escúchame} outperforms state-of-the-art models in all tested configurations by a consequent margin. Indeed, it gains for the active listening and socratic method, respectively: $3.05$ and $9.28$ points against GPT-3.5, $3.25$ and $10.67$ points against LLaMA-2-7b-Chat, and $6.15$ and $15.52$ points against the most recent LLM Mixtral 8x7b. Further analysis is given below.

    \begin{table*}[!ht]
        \centering
        \small
        \resizebox{0.7\textwidth}{!}{
        \begin{tabular}{lcc}
        \hline
        \textbf{Model} & \textbf{Active Listening}& \textbf{Socratic Method}\\
        \hline
        GPT2-124M~\cite{Minghao2023_lamini} & $32.57$& $30.68$\\ 
        Mixtral 8x7b~\cite{jiang2024_mixtral} & $84.52$& $61.60$\\
        LLaMA-2-7b-Chat \cite{Touvron2023_LLaMa2}  & $87.42$& $66.45$\\
        GPT-3.5~\cite{chatGPT3.5} & $87.62$& $67.84$\\
        \textit{Sólo Escúchame} (ours) & $\textbf{90.67}$ & $\textbf{77.12}$ \\
        \hline
        \end{tabular}
        }
        \caption{Final Scores for Psychological Accompaniment Evaluation in Language Models (LMs)}
        \label{tab:results}
    \end{table*}

    Table~\ref{tab:active_listening_results} and Table~\ref{tab:socratic_results} present detailed results obtained with the Active Listening and the Socratic methods, respectively. In both tables, we showcase the accuracy of each criterion by displaying the number of correctly classified samples. This approach allows for a more thorough understanding of the distinct strengths exhibited by each model.

\input{detailed_results}

    \paragraph{GPT-3.5.}

    This model excels in active listening but encounters challenges when employing the Socratic method. For instance, the model presents limitations in generating questions that are designed to cause cognitive dissonance. 
    Contextually and structurally, the responses are adequate, with an average of $77$ tokens, effectively fulfilling the task.
    
    \paragraph{LLaMA-2-7b-Chat.}

    The model exhibits language inconsistency, using English in $8$ of $20$ responses despite Spanish instruction. Responses are lengthy (average $111$ tokens), and frequently initiates unprompted greetings, concluding conversations with farewell-like encouragement, hindering in-depth engagement. 

    \paragraph{GPT2-124M.}

    This baseline performs the worst and may not be suitable for the task. Despite Spanish fine-tuning. The model deforms names and introduces non-existent words. The coherence presented at beginning of the sentence fades away after $50$ tokens. Average response length is $100$ tokens.  

    \paragraph{Mixtral 8x7b.}
    
    Mixtral 8x7b outperforms GPT-3.5 in active listening, but lags in the Socratic method. Despite being over $10$ times smaller, it performs exceptionally well, requiring only a satisfactory prompt and yielding responses with an average length of $78$ tokens.

    \paragraph{Sólo Escúchame.}

    Our model outperforms baselines in both methodologies. It leads in empathy with a score of $291$, surpassing Mixtral. Consistently non-judgmental, well-structured responses with an average length of $78$ tokens. While not perfect and shares limited empathy vocabulary with GPT-3.5, it satisfactorily fulfills the task.

    \subsection{Conversation Example}

    Table~\ref{tab:example_conversation} illustrates a sample conversation between the user and the chatbot. It showcases the coherence of responses, empathetic interactions, and the respect the chatbot demonstrates for the user's expressed feelings. 

     \begin{table}[ht]
        \centering
        \begin{tabular}{|p{1.0\linewidth}|}
            \hline
            \small
             \textbf{User: } Hola, me siento molesta el día de hoy

             \textbf{Chatbot: } Hola Jannette, lamento que te sientas molesta. Si deseas compartir más sobre lo que sucedió, estoy aquí para escucharte.

             \textbf{User: } En serio que como me caga que mi novio haga las cosas y no me las cuente, me las oculte.

             \textbf{Chatbot: } Jannette, es comprensible que te sientas molesta cuando alguien no te comunica lo que está sucediendo en su vida. ¿Quieres hablar más sobre lo que está sucediendo con tu novio y cómo te hace sentir?

             \textbf{User: } Es que siento que no confía en mí para contármelo, y eso me hace sentir excluida.

             \textbf{Chatbot: } Es comprensible que te sientas excluida cuando alguien no te comunica lo que está sucediendo en su vida. La comunicación abierta y honesta es clave para un buen enfoque en una relación. ¿Has intentado hablar con tu novio sobre cómo te sientes?

             \textbf{User: } ¿crees que él lo pueda entender?

             \textbf{Chatbot: } Claro, Jannette. Entiendo que la comunicación abierta y honesta es clave en una relación. Si no has hablado con tu novio sobre cómo te sientes, tal vez sea una buena idea hacerlo. Recuerda que es importante expresar tus sentimientos y necesidades para mantener una relación saludable. ¿Quieres hablar más sobre cómo te sientes?

             \textbf{User: } Creo que no, bueno, no al menos hasta hablar con el y saber su reaccion. Pero gracias

             \textbf{Chatbot: } ¡No hay problema! Estoy aquí para escucharte y brindarte apoyo.
             \\
             \hline
        \end{tabular}
        \caption{Chatbot Interaction Snapshot}
        \label{tab:example_conversation}
    \end{table}

    \section{Conclusions}

    Our model, \textit{Sólo Escúchame}, outperforms the baseline models in both evaluation methodologies. It demonstrates empathy in responses, avoids judgment of user feelings, and maintains vigilance over user-expressed details throughout the conversation. \textit{Sólo Escúchame} is a promising psychological assistant that can be installed locally, running on a CPU and providing flexibility for installation on various machines.
    
    Furthermore, HEAR is the dataset used to train Llama2-7b-chat for emotional accompaniment. This dataset is completely in Spanish, giving a new perspective of how to improve the performance on emotional accompaniment of models with similar capabilities as Llama2-7b-chat in this language. This dataset can highlight the capabilities of language to be more empathetic with users, taking into account their contexts and providing more appropriate attention to the situation.

    \subsection{Limitations and Future Work}
    
    \textit{Sólo Escúchame} is designed to be a supplementary tool for psychologists, offering accessible, non-judgmental support, particularly when professional services are unavailable. The model is still under research. In the near future, we plan to record conversations, allowing individuals to use them later in therapy if they wish. These recordings can serve as reminders of the feelings experienced during moments of anxiety or depression. The authors chose automatic translation to ensure accurate Spanish text and to enhance the model's performance in this language. While automatic translation can introduce some limitations, we thoroughly review generated conversations to ensure their relevance and pertinence. We are committed to improving the translation process and increasing the dataset to enhance the model's performance.

    The HEAR dataset, while sufficient for the model's for emotional accompaniment, limitations in the data prevent achieving this goal fully. Expanding the dataset with a larger and more diverse sample of responses would be beneficial. This would include a broader range of emotions typically found in conversations, allowing the dataset to train models for more extended and complex emotional scenarios.
    
    Additionally, future research  aims to integrate voice input into the system, allowing users to express emotions through text or spoken interactions. The goal is to enhance the system's ability to understand non-verbal cues, discerning deeper emotions beyond explicit words.


\bibliographystyle{splncs04}
\bibliography{custom}




\end{document}

%% file: detailed_results.tex
\begin{table}[!htb]
    \begin{minipage}{.49\linewidth}
      \centering
\begin{tabular}{| l | c | c | c | c | c |}
\cline{2-6}
\multicolumn{1}{c|}{} & \rotatebox{90}{GPT-3.5}  & \rotatebox{90}{LLaMA-2-7b-Chat} & \rotatebox{90}{Mixtral8x7b} & \rotatebox{90}{GPT-2-124M} & \rotatebox{90}{Solo Escúchame}\\
\hline
\makecell{\textbf{Contextual}\\ \textbf{Attention}} & $1256$  & $1260$ & $1277$ & $462$ & $\textbf{1240}$ \\
\hline
\makecell{\textbf{Formulation of}\\ \textbf{Clarifying Questions}} & $776$  & $718$ & $531$ & $199$ & $\textbf{913}$ \\
\hline
\makecell{\textbf{Go Deeper}\\ \textbf{into Conversation}}  & $1215$  & $1240$ & $1185$ & $470$ & $\textbf{1254}$ \\
\hline
\makecell{\textbf{Absence of Judgment}\\ \textbf{or Criticism}}  & $1292$  & $1278$ & $1299$ & $517$ & $\textbf{1300}$\\
\hline
\makecell{\textbf{Demonstration} \\ \textbf{of Empathy}} & $1246$  & $1274$ & $1287$ & $502$ & $\textbf{1278}$\\
\hline
\end{tabular}
\caption{Active listening results. Best in bold.}
\label{tab:active_listening_results}
\end{minipage}
\begin{minipage}{.5\linewidth}
      \centering
\begin{tabular}{| l | c | c | c | c | c |}
\cline{2-6}
\multicolumn{1}{c|}{} & \rotatebox{90}{GPT-3.5}  & \rotatebox{90}{LLaMA-2-7b-Chat} & \rotatebox{90}{Mixtral8x7b} & \rotatebox{90}{GPT-2-124M} & \rotatebox{90}{Solo Escúchame} \\
\hline
\makecell{\textbf{Use of Inductive}\\ \textbf{Questions}} & $1077$  & $1033$ & $872$ & $502$ & $\textbf{1224}$ \\
\hline
\makecell{\textbf{Non-Imposition}\\ \textbf{of Ideas}} & $1236$ & $1170$ & $1200$ & $536$ & $\textbf{1299}$ \\
\hline
\makecell{\textbf{Expansion and Construction}\\ \textbf{of Knowledge}}  & $1031$ & $1071$ & $972$ & $473$ & $\textbf{1245}$ \\
\hline
\makecell{\textbf{Generation of}\\ \textbf{Cognitive Dissonance}}  & $45$  & $36$ & $34$ & $16$ & $\textbf{69}$\\
\hline
\makecell{\textbf{Guided} \\ \textbf{Discovery}} & $1089$ & $1076$ & $988$ & $498$ & $\textbf{1253}$ \\
\hline
\end{tabular}
\caption{Socratic method results. Best in bold.}
\label{tab:socratic_results}
    \end{minipage} 
\end{table}